\documentclass[fleqn,10pt]{wlscirep}
\usepackage[utf8]{inputenc}
\usepackage[T1]{fontenc}
\usepackage{xcolor}
\usepackage{tabularx}
\usepackage{multirow}
\title{Continuous wavelet-based feature extraction for
laser-induced breakdown spectroscopy classification
of human plasma samples}

\author[1]{Md Mahmudur Rahman}
\author[1]{Atqiya Munawara Mahi}
\author[1]{Nazim Ahmed Belabbaci}
\author[1,2,*]{Mohammad Arif Ul Alam}
\affil[1]{Miner School of Computer and Information Sciences, Kennedy College of Sciences, University of Massachusetts, Lowell, USA}
\affil[2]{Department of Computer Systems Technology, North Carolina Agriculture and Technical University}

\affil[*]{malam7@ncat.edu}


\begin{abstract}
Extracting discriminative features from laser-induced breakdown spectroscopy (LIBS) spectra is important for the classification of complex biological samples. This study evaluates continuous wavelet transform (CWT)-based spectral peak extraction for distinguishing human plasma samples labeled as SARS-CoV-2-positive or negative. Plasma specimens from 181 individuals, comprising 53 positive samples and 128 negative controls, were deposited as 5~$\mu$L droplets on silicon substrates and dried before analysis. Spectra were acquired using a nanosecond Nd:YAG laser operating at 1064~nm, with 100 single-shot measurements collected per specimen and averaged after intensity-based outlier rejection. CWT-based peak detection was coupled with a support vector machine (SVM) classifier. The reported leave-one-out evaluation yielded an accuracy of 93.00\%, sensitivity of 94.67\%, and specificity of 91.61\%; these estimates remain provisional pending reconciliation with specimen-level predictions. Features identified by the CWT pipeline included emission lines assigned to Ca~II, K~I, and N~I. The workflow provides a basis for examining spectral features associated with the studied plasma groups. Independent validation and assessment of specimen, substrate, and acquisition-related variability are required to establish generalizability and diagnostic relevance.
\end{abstract}

\begin{document}

\flushbottom
\maketitle

\thispagestyle{empty}

\section{Introduction}

Laser-induced breakdown spectroscopy (LIBS) offers a multielemental approach to characterizing biological specimens through their atomic and ionic emission spectra. A focused laser pulse ablates a small amount of material and generates a transient plasma, whose emission provides information about the elemental composition of the interrogated sample \cite{TOGNONI20021115}. The ability to acquire multiple elemental signatures from small sample volumes, with comparatively limited chemical preparation, has motivated applications in biological and biomedical analysis \cite{GAUDIUSO2019123}. For complex biofluids, however, the analytical value of LIBS depends on more than detecting emission lines: it also requires spectral representations that preserve useful sample differences while limiting the influence of background emission, measurement variability, and matrix effects.

Blood plasma is a particularly challenging matrix because its spectra reflect contributions from endogenous constituents, specimen preparation, and the conditions under which the laser-induced plasma is formed. Emission intensity cannot generally be interpreted as elemental concentration without accounting for excitation conditions, ablation behavior, and matrix-dependent effects \cite{TOGNONI20021115,GAUDIUSO2019123}. Drying a specimen on a solid substrate further introduces potential variability associated with deposition morphology and substrate emission. Consequently, identifying spectral features associated with sample labels requires careful distinction between biological variation and variation introduced during preparation or acquisition. This distinction is especially relevant when classification models are developed from relatively small collections of biological specimens.

Previous studies have established the feasibility of combining LIBS with statistical learning for biological sample discrimination. Applications include the classification of protein biomarker solutions \cite{pokrajac2014automatic}, discrimination of plasma samples from healthy and ovarian cancer-prone mice \cite{melikechi2016age}, and investigations of Alzheimer's disease using LIBS and machine learning \cite{GAUDIUSO2020105931}. These studies motivate the use of multivariate spectral patterns in biomedical analysis, while also emphasizing the need to establish the analytical basis and generalizability of observed discrimination. A spectral feature associated with a disease-labeled group is not, by itself, evidence of a disease-specific biomarker.

SARS-CoV-2-associated plasma samples provide a relevant setting for investigating these methodological questions. Molecular and antigen tests assess viral targets, whereas LIBS-based analysis of plasma examines spectral patterns that may reflect host-associated compositional differences. These approaches therefore address different analytical signals. Berlo et al.\ demonstrated that LIBS combined with machine learning could distinguish plasma from donors who had previously tested positive for SARS-CoV-2 by reverse-transcription polymerase chain reaction from control plasma \cite{berlo2022laser}. Their findings establish an important precedent for the present application. Further methodological evaluation should therefore address how spectral information is represented and used for classification, rather than assume that classification performance alone establishes direct viral detection or clinical diagnostic validity.

Feature extraction is central to this problem. Full LIBS spectra contain numerous correlated variables, while potentially informative differences may be concentrated within a smaller set of emission features. With limited numbers of independent specimens, the choice of spectral representation can influence both classification performance and the interpretation of the resulting model. Continuous wavelet transform (CWT)-based peak detection uses localized responses across spectral scales to identify candidate peaks; wavelet-based methods have previously been investigated for analytical peak detection \cite{du2006improved}. The resulting spectral representation depends on peak structure, background variation, and peak-detection settings.

In this study, we evaluate CWT-based peak detection for the classification of human plasma samples labeled as SARS-CoV-2-positive or negative. The dataset comprises 181 individual specimens, including 53 positive samples and 128 negative controls. Spectra were acquired from dried plasma droplets on silicon substrates, and the extracted peak features were used to train an SVM classifier. The evaluation focuses on classification performance under the reported leave-one-out cross-validation procedure, the emission features retained by the peak-based representation, and the effects of selected wavelength exclusions and wavelet settings.

The contribution of this work is an evaluation of CWT-based
spectral peak extraction within a specimen-level LIBS
classification workflow. Using plasma specimens labeled as
SARS-CoV-2-positive or negative, we examine the resulting
classification performance, the sensitivity of peak selection
to wavelet settings, and the effects of selected wavelength
exclusions. The analysis is intended to characterize the
behavior of the spectral representation within the available
specimen collection; independent validation is required to
establish its generalizability.

\begin{figure}[htbp]
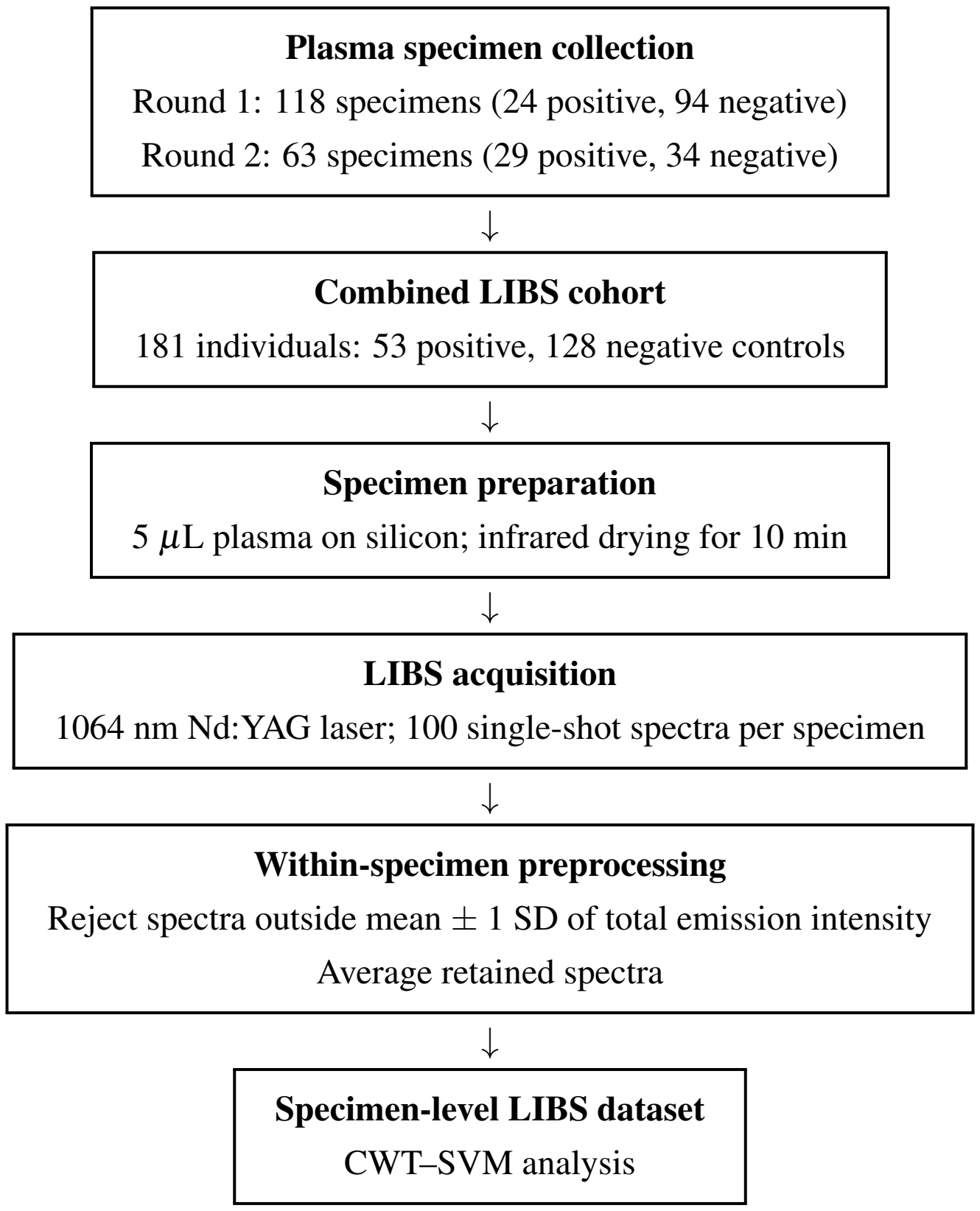

\centering
\begingroup
\small
\setlength{\tabcolsep}{7pt}
\renewcommand{\arraystretch}{1.25}
\begin{tabular}{c}
\fbox{\begin{tabular}{c}
\textbf{Plasma specimen collection}\\
Round 1: 118 specimens (24 positive, 94 negative)\\
Round 2: 63 specimens (29 positive, 34 negative)
\end{tabular}}\\
$\downarrow$\\
\fbox{\begin{tabular}{c}
\textbf{Combined LIBS cohort}\\
181 individuals: 53 positive, 128 negative controls
\end{tabular}}\\
$\downarrow$\\
\fbox{\begin{tabular}{c}
\textbf{Specimen preparation}\\
5~$\mu$L plasma on silicon; infrared drying for 10 min
\end{tabular}}\\
$\downarrow$\\
\fbox{\begin{tabular}{c}
\textbf{LIBS acquisition}\\
1064 nm Nd:YAG laser; 100 single-shot spectra per specimen
\end{tabular}}\\
$\downarrow$\\
\fbox{\begin{tabular}{c}
\textbf{Within-specimen preprocessing}\\
Reject spectra outside mean $\pm$ 1 SD of total emission intensity\\
Average retained spectra
\end{tabular}}\\
$\downarrow$\\
\fbox{\begin{tabular}{c}
\textbf{Specimen-level LIBS dataset}\\
CWT--SVM analysis
\end{tabular}}
\end{tabular}
\endgroup
\caption{Plasma specimen collection, preparation, and LIBS data acquisition. Two collection rounds yielded 181 specimens, comprising 53 SARS-CoV-2-positive samples and 128 negative controls. Following deposition and drying on silicon substrates, 100 single-shot spectra were acquired per specimen. Spectra with total emission intensities outside the within-specimen mean $\pm$ one standard deviation were excluded, and retained spectra were averaged for subsequent CWT-SVM analysis. Sample counts refer to individual specimens rather than laser shots.}
\label{fig:libs_sample_workflow}
\end{figure}

\section{Datasets}

\subsection{Blood Plasma LIBS Spectra Sample Collection and Initial Signal Processing}

A total of 181 human plasma specimens were collected in sodium citrate anticoagulant from donors drawn by MRN Diagnostics both before (pre 11/2019) and after the appearance of SARS-CoV-2 in the United States. Informed consent was obtained from all subjects and studies were performed in accordance with FDA guidelines and regulations including IRB approval. Specimens were collected in two rounds: in the first round, 118 samples were collected (24 SARS-CoV-2 `positive' and 94 `negative' control) and in the second round, 63 samples were collected (29 SARS-CoV-2 `positive' and 34 `negative' control) samples. The 181 samples provided by MRN Diagnostics were analyzed by University of Massachusetts using LIBS technology.

Prior to performing LIBS measurements of the blood plasma samples, 5~$\mu$L of each individual blood plasma specimen were deposited on the unpolished side of pure $Si$ wafers. These wafers were previously rinsed in 2-propanol. The blood plasma samples deposited on the $Si$ wafers were then dried for 10 min using a Tungsten infrared lamp. During the laser ablation process, separate single-shot spectra were captured from adjacent spots within the deposited droplet.

The experimental setup for LIBS involved focusing 7 ns Nd:YAG laser pulses (Surelite II, Continuum) with a wavelength of 1064 nm onto the samples using an air-spaced doublet lens with a focal length of 30 mm. The blood plasma specimens were loaded onto a computer-controlled translation stage within a SciTrace chamber (AtomTrace), and the emission from the laser-induced plasma was collected using a 50 $\mu$m core-diameter optical fiber positioned at a $45^\circ$ angle relative to the laser beam. The optical fiber was connected to an Echelle spectrograph (Andor Technology, ME 5000) and a thermoelectrically cooled iStar Intensified Charge Coupled Device (ICCD) camera (Andor Technology, DH734-18F-03). The time parameters utilized for this work were as follows: a gate delay of 1 microsecond and a gate width of 5 microseconds. The focused laser spot had a diameter of approximately 100 $\mu$m, operating at a repetition rate of $1$ Hz with a laser energy of 130$\pm$2 mJ. All measurements were conducted in ambient air pressure. For each blood plasma sample, a total of 100 single-shot spectra were collected. Each spectrum was obtained from a new location on the dried blood plasma drop by using the 3-D translation stage after each laser shot. During analysis, the 100 individual spectra obtained for each sample were averaged, with any outlier spectra - those falling outside the range of mean $\pm 1$ standard deviation of total emission intensity - being excluded from the average process.

\begin{figure}[htbp]
\centering
\includegraphics[width=0.9\linewidth]{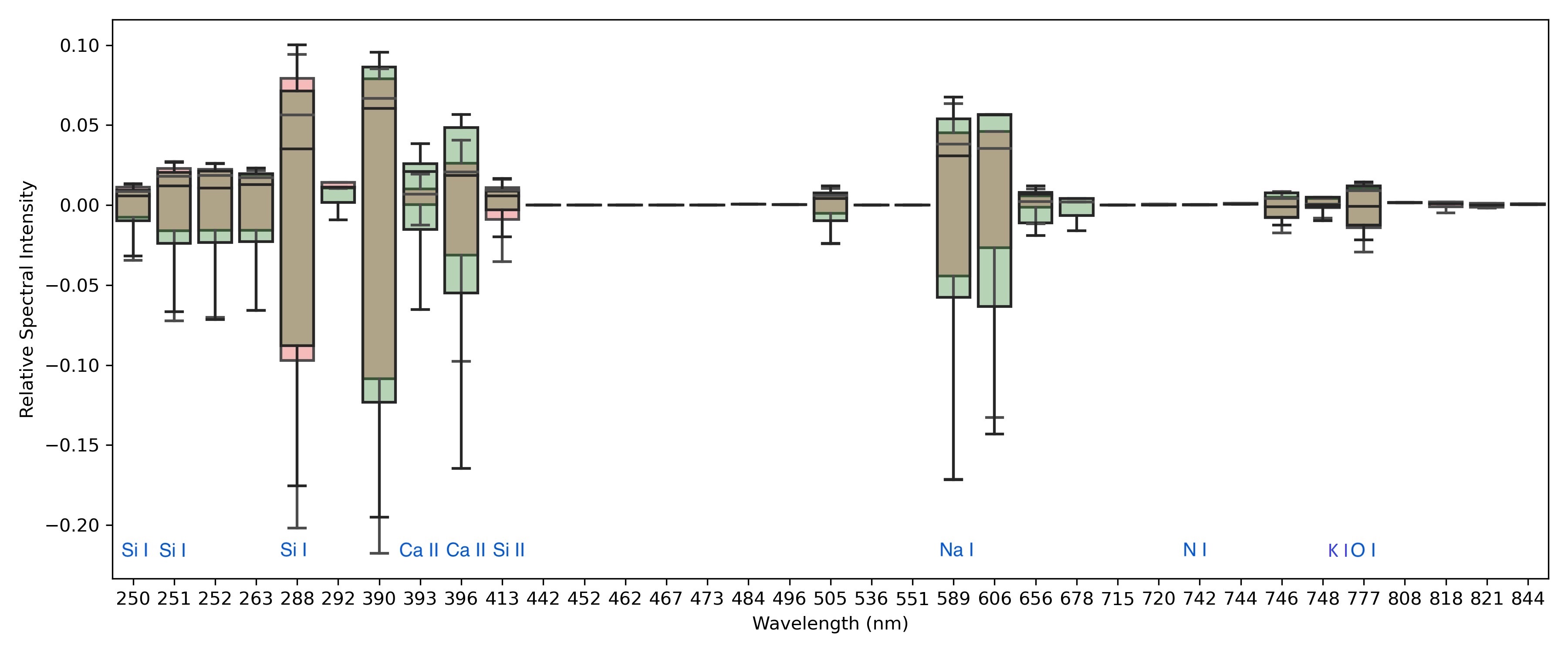}
\caption{Median specimen-level LIBS intensities in the positive (green) and negative (red) groups at selected wavelengths. Error bars indicate the first and third quartiles. The horizontal axis shows wavelength in nanometers.}
\label{fig:comp}
\end{figure}

\section{Methods}
\label{sec:methods}

\subsection{Analytical workflow and unit of analysis}
\label{subsec:analytical_workflow}

The analysis combined continuous wavelet transform (CWT)-based
spectral peak detection with support vector machine (SVM)
classification. Specimen preparation, LIBS acquisition, and
intensity-based rejection of individual laser-shot spectra are
described in the preceding section. Following these procedures,
the retained spectra were averaged within each specimen to obtain
one representative spectrum per individual. The individual
specimen, rather than the laser shot, was the unit of analysis.

The resulting dataset comprised 181 specimen-level spectra,
including 53 specimens labeled as SARS-CoV-2-positive and
128 negative controls. The analytical workflow consisted of
detecting candidate emission peaks, aligning their wavelength
positions across specimens, constructing peak-intensity feature
vectors, and classifying the specimens using an SVM. Classification
was evaluated against the supplied specimen labels.

\subsection{Continuous wavelet transform for spectral peak detection}
\label{subsec:cwt_detection}

CWT-based peak detection was used to identify localized features
in the specimen-level LIBS spectra. Wavelet methods characterize
spectral structure by comparing the measured signal with translated
and scaled versions of a reference function, providing a basis
for detecting peaks over a range of characteristic widths
\cite{du2006improved,stephane1999wavelet}.

Let $f_i(t)$ denote the averaged emission intensity of specimen
$i$ at spectral coordinate $t$. Its continuous wavelet transform
is expressed as
\begin{equation}
    W_i(u,s)
    =
    \frac{1}{\sqrt{s}}
    \int_{-\infty}^{+\infty}
    f_i(t)
    \psi^{*}\left(\frac{t-u}{s}\right)
    \,\mathrm{d}t,
    \qquad s>0,
    \label{eq:cwt_transform}
\end{equation}
where $u$ is the translation parameter, $s$ is the scale parameter,
$\psi$ is the mother wavelet, and the asterisk denotes complex
conjugation. The transform was applied to the discretely sampled
spectra. Spectral coordinates were associated with their measured
wavelengths for subsequent peak alignment and interpretation.

A Mexican-hat wavelet was used because its central positive lobe
and flanking negative lobes provide a localized representation of
peak-like structure. A normalized form is
\begin{equation}
    \psi(t)
    =
    \frac{2}{\sqrt{3}\,\pi^{1/4}}
    \left(1-t^{2}\right)
    \exp\left(-\frac{t^{2}}{2}\right).
    \label{eq:mexican_hat}
\end{equation}
This real-valued wavelet has zero mean and is proportional to the
negative second derivative of a Gaussian. Translation determines
the position at which the spectrum is examined, whereas scale
controls the characteristic width of the structure to which the
wavelet responds.

\begin{figure}[htbp]
    \centering
    \includegraphics[width=0.95\linewidth]{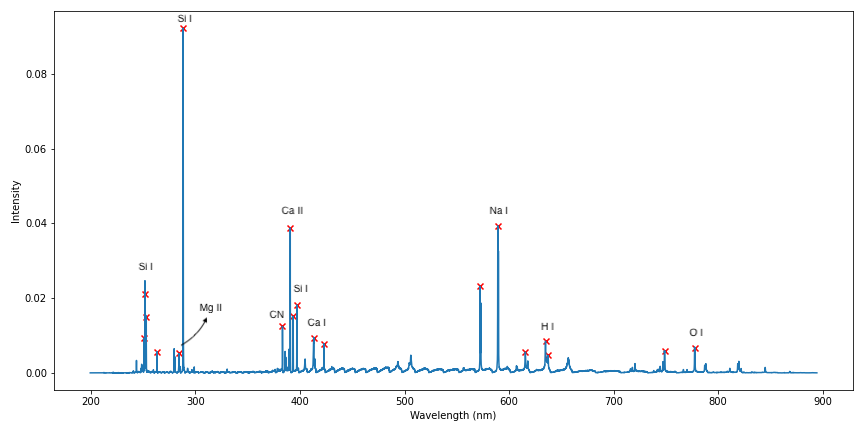}
    \caption{Illustration of CWT-based peak detection in a
    specimen-level LIBS spectrum. The blue curve represents
    the measured spectrum, and red markers indicate retained
    peak positions.}
    \label{fig:peak_detection}
\end{figure}

Candidate peaks were identified from the wavelet response and
screened using a signal-to-noise criterion. The retained peak
positions were then associated with their intensities in the
specimen-level spectrum. Figure~\ref{fig:peak_detection}
illustrates the resulting peak detections.

\subsection{Peak alignment and feature construction}
\label{subsec:feature_construction}

Detected peaks were aligned by wavelength to establish
corresponding spectral features across specimens. This alignment
provided a common feature representation in which each column
corresponded to a matched peak position and each row corresponded
to an individual specimen. For $p$ aligned features, the feature
vector for specimen $i$ was
\begin{equation}
    \mathbf{x}_i
    =
    \left(x_{i1},x_{i2},\ldots,x_{ip}\right)^{\mathsf{T}},
    \label{eq:feature_vector}
\end{equation}
where $x_{ij}$ denotes the spectral peak height assigned to
feature $j$ in specimen $i$. The complete feature matrix was
$\mathbf{X}\in\mathbb{R}^{N\times p}$, with $N=181$.
The response variable was $y_i=1$ for a positive specimen and
$y_i=0$ for a negative control.

Peak heights were treated as empirical spectral descriptors.
They were not converted to elemental concentrations, and their
association with specimen labels was not interpreted as evidence
of disease-specific elemental biomarkers.

\subsection{Support vector machine classification}
\label{subsec:svm_classification}

An SVM was used to classify specimens from their aligned
peak-intensity vectors \cite{noble2006support,pisner2020support}.
The reported analysis examined linear and nonlinear kernels and
selected a radial basis function (RBF) kernel. For feature
vectors $\mathbf{x}_i$ and $\mathbf{x}_j$, the RBF kernel is
\begin{equation}
    K(\mathbf{x}_i,\mathbf{x}_j)
    =
    \exp\left(
    -\gamma
    \left\|\mathbf{x}_i-\mathbf{x}_j\right\|_2^{2}
    \right),
    \label{eq:rbf_kernel}
\end{equation}
where $\gamma>0$ determines the kernel width. The regularization
parameter $C$ controls the trade-off between the margin penalty
and classification errors. The reported CWT-SVM configuration
used $C=10^{-3}$.

\subsection{Cross-validation and performance measures}
\label{subsec:validation}

The reported classification analysis used leave-one-out
cross-validation at the specimen level. In each iteration, one
specimen was withheld for evaluation and the remaining 180
specimens were used for classifier training. This procedure was
repeated until every specimen had served once as the evaluation
case. The procedure therefore assessed internal predictive
performance within the available specimen collection.


For evaluation, positive specimens were designated as the positive
class. Classification performance is defined from the pooled
out-of-fold class predictions as
\begin{align}
    \mathrm{Accuracy}
    &=
    \frac{TP+TN}{TP+TN+FP+FN},
    \label{eq:accuracy}\\
    \mathrm{Sensitivity}
    &=
    \frac{TP}{TP+FN},
    \label{eq:sensitivity}\\
    \mathrm{Specificity}
    &=
    \frac{TN}{TN+FP},
    \label{eq:specificity}
\end{align}
where $TP$, $TN$, $FP$, and $FN$ denote the numbers of true-positive,
true-negative, false-positive, and false-negative predictions,
respectively. Values are expressed as percentages in the results.

\subsection{Peak-detection settings and wavelength-exclusion analysis}
\label{subsec:sensitivity_analysis}

\begin{figure}[htbp]
    \centering
    \includegraphics[width=0.95\linewidth]
    {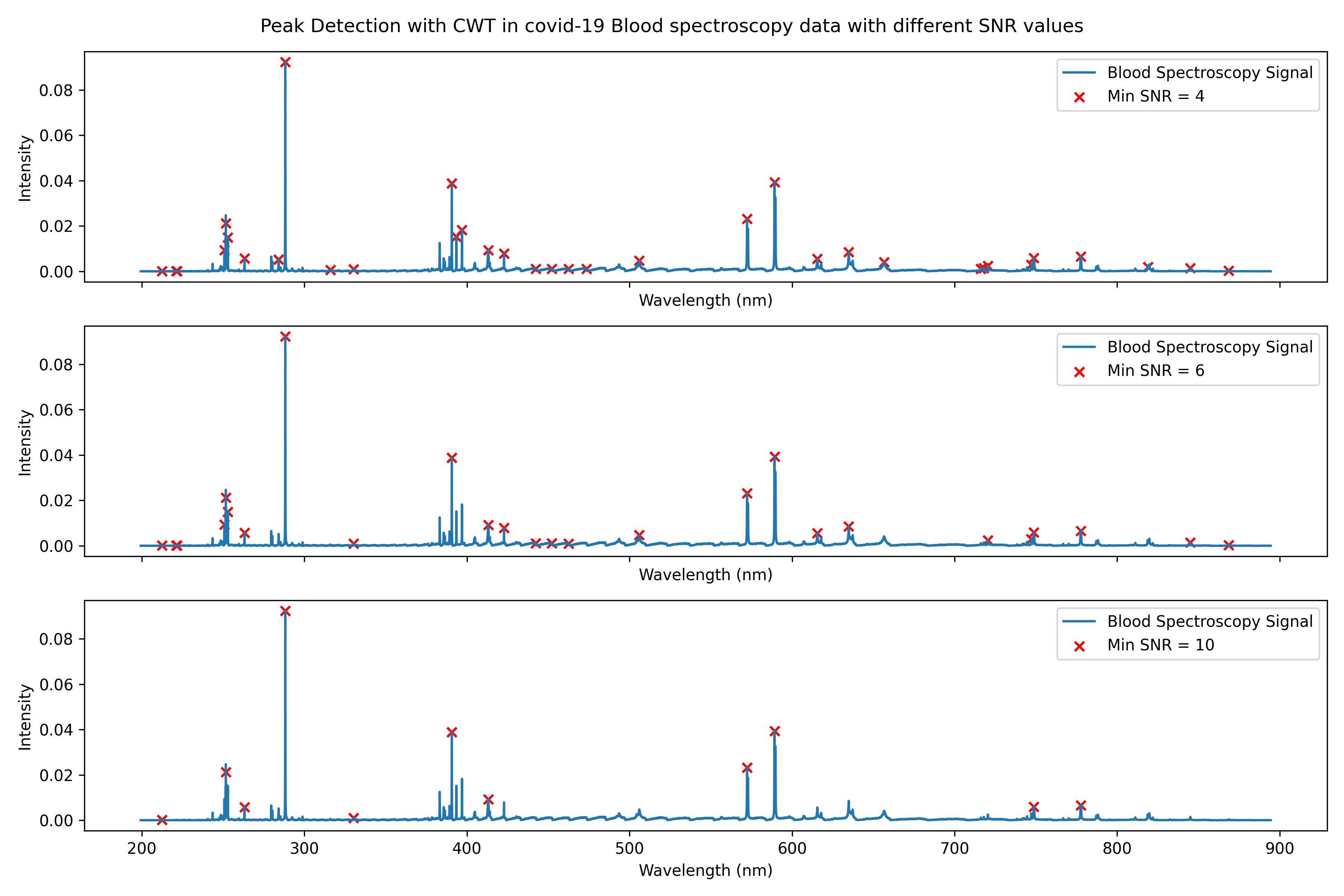}
    \caption{Effect of the signal-to-noise setting on CWT-based
    peak detection in a LIBS spectrum. The comparison illustrates
    changes in the set of retained spectral peaks.}
    \label{fig:peak_detection_snr}
\end{figure}

The influence of peak-detection settings was examined by varying
the signal-to-noise criterion and wavelet width. The resulting
peak selections were compared to assess how these parameters
affected the retention of weak features and the separation of
neighboring peaks. Figure~\ref{fig:peak_detection_snr} illustrates
the effect of the signal-to-noise setting, and
Figure~\ref{fig:peak_detection_width} illustrates the effect of
wavelet width. The reported analysis used a wavelet-width
parameter of 20; this value denotes an algorithm parameter
rather than a physical emission-line width unless a conversion
to wavelength units is specified.


A wavelength-exclusion analysis was also conducted for the
CWT-SVM pipeline. Features corresponding to selected wavelengths
were omitted individually, and classification performance was
compared with that obtained using the complete feature set.
The reported exclusions concerned features near 396.880,
288.179, 413.085, and 393.398~nm. These comparisons assess the
dependence of the classifier on particular spectral features;
they do not establish a causal relationship between an element
and the specimen label.

\section{Results}
\label{sec:results}

\subsection{Specimen-level classification}
\label{subsec:classification_results}

The analytical dataset comprised 181 plasma specimens, including
53 SARS-CoV-2-positive specimens and 128 negative controls.
The retained classification pipeline used CWT-derived spectral
peak intensities as inputs to an SVM with an RBF kernel.
The existing evaluation reported an accuracy of 93.00\%,
sensitivity of 94.67\%, and specificity of 91.61\%
(Table~\ref{tab:result}). These estimates describe the reported
internal evaluation of the available specimen collection.

\begin{table}[htbp]
    \centering
    \caption{Reported classification performance of the
    CWT-SVM pipeline. Values are reproduced from the existing
    analysis and remain subject to verification against the
    pooled out-of-fold predictions.}
    \label{tab:result}
    \small
    \renewcommand{\arraystretch}{1.15}
    \begin{tabular}{lccc}
        \hline
        Method &
        Accuracy (\%) &
        Sensitivity (\%) &
        Specificity (\%) \\
        \hline
        CWT-SVM & 93.00 & 94.67 & 91.61 \\
        \hline
    \end{tabular}
\end{table}

\subsection{Spectral features and peak-detection behavior}
\label{subsec:spectral_results}

The reported CWT analysis highlighted spectral features at
393.398, 396.880, 742.512, and 766.49~nm. The corresponding
assignments in the original analysis were Ca~II, Ca~II, N~I,
and K~I, respectively. These assignments describe candidate
emission features and do not establish their specificity to
SARS-CoV-2-associated specimens.

\begin{figure}[htbp]
\centering
\includegraphics[width=0.99\linewidth]{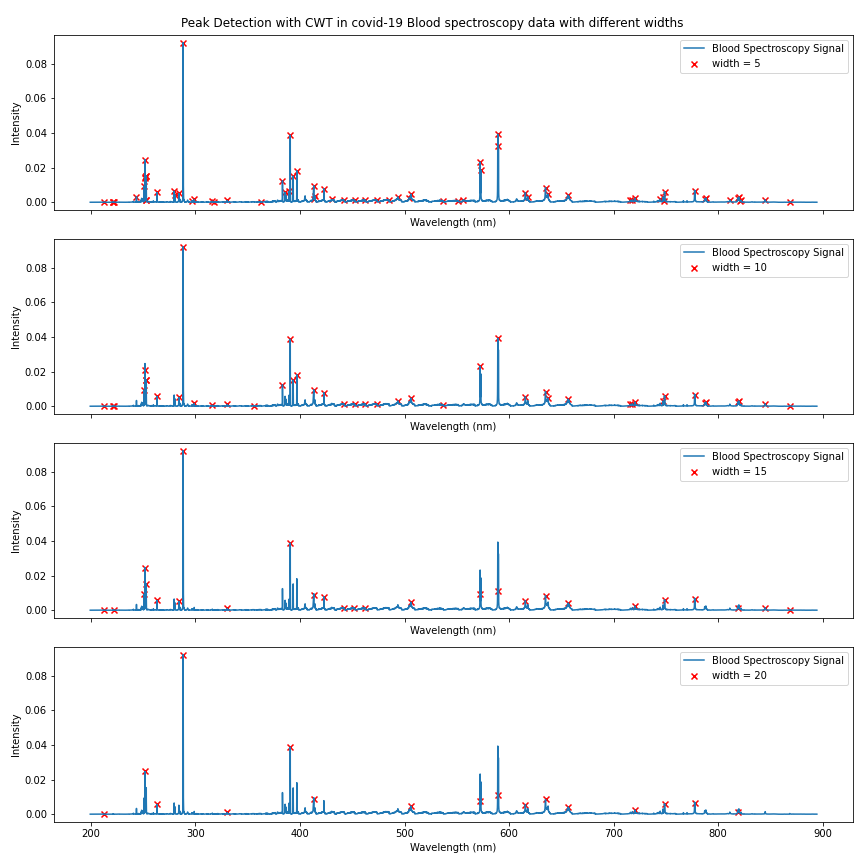}
\caption{Effect of the wavelet-width parameter on CWT-based
peak detection in a representative LIBS spectrum. Different
settings alter the number and positions of retained candidate
peaks. Width values denote algorithm parameters.}
\label{fig:peak_detection_width}
\end{figure}

The peak-detection examples illustrate the dependence of the
extracted representation on the wavelet settings.
The original qualitative comparison indicated that smaller
wavelet widths retained more local spectral features, whereas
larger widths omitted some narrower features
(Figure~\ref{fig:peak_detection_width}).
The reported analysis adopted a width parameter of 20.
The signal-to-noise comparison illustrates how the retained
peak set changed with the detection criterion
(Figure~\ref{fig:peak_detection_snr}).
These comparisons characterize peak-selection behavior;
quantitative performance differences between the tested
settings were not reported.

\subsection{Wavelength-exclusion analysis}
\label{subsec:wavelength_exclusion_results}

The reported wavelength-exclusion results are presented in
Table~\ref{tab:ablation}. Relative to the reported complete-feature
accuracy of 93.00\%, exclusion of the feature at 396.880~nm
yielded an accuracy of 92.42\%, a decrease of
0.58 percentage points. Exclusion of the features at
288.179 and 413.085~nm yielded accuracies of 87.45\%
and 88.23\%, corresponding to decreases of 5.55 and
4.77 percentage points, respectively. The largest reported
accuracy decrease occurred after exclusion of the feature
at 393.398~nm, for which accuracy was 84.65\%, or
8.35 percentage points below the complete-feature value.

\begin{table}[htbp]
    \centering
    \caption{Reported wavelength-exclusion results for the
    CWT-SVM pipeline. The accuracy decrease is calculated
    relative to the reported complete-feature accuracy of
    93.00\% and is expressed in percentage points (pp).
    All performance values remain provisional pending
    verification of their calculation and evaluation denominators.}
    \label{tab:ablation}
    \small
    \setlength{\tabcolsep}{5pt}
    \renewcommand{\arraystretch}{1.15}
    \begin{tabular}{lrrrr}
        \hline
        \shortstack[l]{Excluded\\wavelength (nm)} &
        \shortstack{Accuracy\\(\%)} &
        \shortstack{Sensitivity\\(\%)} &
        \shortstack{Specificity\\(\%)} &
        \shortstack{Accuracy\\decrease (pp)} \\
        \hline
        None    & 93.00 & 94.67 & 91.61 & 0.00 \\
        396.880 & 92.42 & 91.65 & 89.46 & 0.58 \\
        288.179 & 87.45 & 82.34 & 84.87 & 5.55 \\
        413.085 & 88.23 & 84.67 & 83.65 & 4.77 \\
        393.398 & 84.65 & 82.13 & 80.91 & 8.35 \\
        \hline
    \end{tabular}
\end{table}

These numerical differences summarize the existing analysis;
their statistical significance has not been established.
In particular, the reported decrease following exclusion of
288.179~nm suggests a possible dependence on this feature, subject to
verification of the performance calculations.
Interpretation of the exclusion results requires verification
of the performance estimates and confirmation of the corresponding
spectral assignments.

\section{Discussion}
\label{sec:discussion}

\subsection{Analytical significance of the CWT-SVM workflow}

This study examined CWT-based spectral feature extraction coupled
with SVM classification for the analysis of 181 human plasma
specimens. The analytical contribution is the use of localized
spectral peaks to represent complex LIBS measurements in a
specimen-level classification workflow. This representation
connects the classifier inputs to identifiable spectral positions,
facilitating examination of the effects of peak-detection settings
and individual wavelength exclusions. The present results should,
however, be interpreted as an exploratory evaluation within the
available specimen collection.

The reported classification estimates suggest that the extracted
features contain information associated with the supplied specimen
labels. Nevertheless, the inconsistencies identified in the
performance summaries prevent firm conclusions about the magnitude
of predictive performance. Verification against the original
out-of-fold predictions is necessary before the reported accuracy,
sensitivity, specificity, or wavelength-exclusion differences can
be interpreted quantitatively. In particular, these estimates
cannot currently establish clinical diagnostic validity or
generalizability to independently collected specimens.

Previous investigations have demonstrated the feasibility of
combining LIBS with statistical learning for protein-solution
classification and biomedical plasma analysis
\cite{pokrajac2014automatic,GAUDIUSO2020105931}.
Berlo et al.\ also reported discrimination of plasma associated
with prior SARS-CoV-2 positivity using LIBS and machine learning
\cite{berlo2022laser}. Accordingly, the present study addresses
spectral representation and feature dependence within this
application rather than establishing the first demonstration of
SARS-CoV-2-associated plasma discrimination. Differences in
specimen composition, acquisition conditions, and validation
procedures preclude direct comparisons of classification
performance across studies.

\subsection{Spectral interpretation and substrate effects}

The CWT analysis highlighted features assigned in the original
analysis to Ca~II, K~I, and N~I. These assignments provide candidate
spectral locations for further examination, but their biological
interpretation remains conditional on wavelength verification,
assessment of spectral overlap, and characterization of
measurement variability. Atomic and ionic emission intensities
depend on both specimen composition and the conditions of
laser ablation and plasma excitation
\cite{TOGNONI20021115,GAUDIUSO2019123}. Consequently, differences
in peak intensity cannot be interpreted directly as differences
in circulating elemental concentrations.

The wavelength-exclusion analysis offers a way to examine the
classifier's dependence on selected features. Conditional on
verification of the reported metrics, the larger decrease
associated with exclusion of the feature near 393.398~nm would
motivate further investigation of that spectral region.
It would not establish a causal contribution of calcium to
SARS-CoV-2 infection or demonstrate that this line is a
disease-specific biomarker. Correlation among spectral features
also limits the interpretation of single-feature exclusions:
other variables may retain related information, while removal
of one feature can alter the fitted decision boundary.

The feature near 288.179~nm requires particular attention because
it was assigned to Si~I in the original analysis and the specimens
were deposited on silicon substrates. The reported exclusion
results suggest a possible dependence on this feature, subject
to verification of the performance calculations. A substrate-associated feature
could reflect differences in specimen coverage, dried-droplet
morphology, or laser interaction with the underlying surface.
These possibilities remain hypotheses rather than demonstrated
mechanisms in the present study. Blank-substrate measurements,
replicate deposits, spatially resolved acquisition, and comparisons
across substrate batches would help distinguish specimen-related
information from preparation-dependent effects.

\subsection{Influence of wavelet settings}

Figure~\ref{fig:peak_detection_width} illustrates the dependence
of the extracted peak set on wavelet width. Smaller widths
respond to finer spectral structure and may retain both narrow
emission features and local fluctuations. Larger widths emphasize
broader structures and may fail to resolve some closely spaced
or narrow peaks. The analytical objective is therefore to identify
settings that preserve reproducible spectral information while
limiting sensitivity to noise.

The width parameter of 20 used in the reported analysis should
be regarded as an implementation-specific choice. The qualitative
comparison does not establish its statistical optimality, and
its physical interpretation depends on the spectral sampling
interval and the definition of width in the implementation.
Similarly, the signal-to-noise criterion influences which weak
features enter the classifier
(Figure~\ref{fig:peak_detection_snr}). Together, these observations
support evaluating peak-detection settings as part of the complete
classification pipeline, rather than selecting them solely
from visual inspection of representative spectra.

Future evaluation should quantify the stability of detected
peaks across replicate measurements and training subsets.
Parameter selection should occur within the training data,
with performance assessed on specimens excluded from that
selection. Such an analysis would clarify whether the observed
discrimination persists across reasonable choices of wavelet
width, signal-to-noise threshold, and wavelength-alignment
tolerance.

\subsection{Study limitations}

The specimen collection is limited in size and contains fewer
positive specimens than negative controls. Although multiple
laser shots were acquired from each deposit, these measurements
are technical replicates and do not increase the number of
independent individuals. Leave-one-out cross-validation provides
an internal evaluation, but it does not establish reproducibility
across collection sites, instruments, or clinical populations.
Its interpretation also depends on whether feature alignment,
scaling, parameter selection, and any other data-dependent
operations were performed independently within the training
portion of each iteration.

The two collection rounds and the inclusion of specimens obtained
before the emergence of SARS-CoV-2 introduce possible temporal
and batch-related confounding. Storage history, specimen handling,
and acquisition conditions could differ between groups and
contribute to apparent discrimination. The available analysis
does not establish that these influences have been excluded.
Evaluation across collection rounds, accompanied by appropriate
specimen metadata and laboratory quality controls, is therefore
needed to assess whether the classifier captures reproducible
sample differences.

Clinical interpretation is additionally constrained by the absence
of demonstrated specificity against other infectious or
inflammatory conditions. A classifier trained on the supplied
positive and negative labels may respond to host-associated
changes or other group differences without detecting the virus
itself. The present analysis consequently does not establish
active-infection detection, performance in asymptomatic
individuals, or transferability across viral variants.

Finally, the current performance summaries require reconciliation
with the specimen-level predictions. Confidence intervals,
a verified confusion matrix, and comparisons with conventional
chemometric baselines would provide a stronger basis for assessing
the CWT-SVM pipeline. The practical advantages of the workflow
also remain to be measured: the use of small plasma volumes and
limited chemical preparation does not by itself demonstrate
short clinical turnaround time, low operating cost, or
point-of-care suitability.

\section{Conclusion and Future Works}
\label{sec:conclusion}

This study presents a CWT-based peak-extraction and SVM
classification workflow for LIBS spectra acquired from
181 human plasma specimens. The workflow represents each specimen
through aligned spectral peak intensities and provides a basis
for examining the effects of peak-detection settings and
selected wavelength exclusions. The current analysis identifies
candidate spectral regions for further investigation, while
verification of the performance calculations remains necessary
before quantitative claims about classification accuracy can
be finalized.

Further work should first establish a reproducible evaluation
using verified specimen-level predictions, complete reporting
of preprocessing and model settings, and training-restricted
parameter selection. Independent specimen collections and
comparisons with established chemometric methods are needed
to assess generalizability and the added value of CWT-based
feature extraction. Analytical validation should include
replicate deposits, substrate blanks, acquisition-day controls,
and assessment of peak stability across instruments and
measurement conditions.

Subsequent clinical studies should include contemporaneous
controls and specimens representing other infectious and
inflammatory conditions, with documented collection and
storage histories. Where feasible, paired elemental measurements
would help evaluate the relationship between LIBS features
and specimen composition. These steps would support a more
rigorous assessment of the workflow as an analytical approach
for plasma characterization and determine whether it warrants
further development for a defined clinical application.

\subsection*{Data Availability Statement}
The data and source code will be available upon reasonable request to the authors.


\begin{thebibliography}{10}

\bibitem{TOGNONI20021115}
E. Tognoni, V. Palleschi, S. Corsi and G. Cristoforetti,
Quantitative micro-analysis by laser-induced breakdown spectroscopy:
a review of the experimental approaches,
\textit{Spectrochimica Acta Part B: Atomic Spectroscopy}
57 (2002) 1115--1130.
doi:10.1016/S0584-8547(02)00053-8.

\bibitem{GAUDIUSO2019123}
R. Gaudiuso et al.,
Laser-induced breakdown spectroscopy for human and animal health:
a review,
\textit{Spectrochimica Acta Part B: Atomic Spectroscopy}
152 (2019) 123--148.
doi:10.1016/j.sab.2018.11.006.

\bibitem{pokrajac2014automatic}
D. Pokrajac et al.,
Automatic classification of laser-induced breakdown spectroscopy
(LIBS) data of protein biomarker solutions,
\textit{Applied Spectroscopy} 68 (2014) 1067--1075.

\bibitem{melikechi2016age}
N. Melikechi et al.,
Age-specific discrimination of blood plasma samples of healthy
and ovarian cancer prone mice using laser-induced breakdown spectroscopy,
\textit{Spectrochimica Acta Part B: Atomic Spectroscopy}
123 (2016) 33--41.

\bibitem{GAUDIUSO2020105931}
R. Gaudiuso, E. Ewusi-Annan, W. Xia and N. Melikechi,
Diagnosis of Alzheimer's disease using laser-induced breakdown
spectroscopy and machine learning,
\textit{Spectrochimica Acta Part B: Atomic Spectroscopy}
171 (2020) 105931.
doi:10.1016/j.sab.2020.105931.

\bibitem{berlo2022laser}
K. Berlo et al.,
Laser induced breakdown spectroscopy for the rapid detection
of SARS-CoV-2 immune response in plasma,
\textit{Scientific Reports} 12 (2022) 1614.
doi:10.1038/s41598-022-05509-z.

\bibitem{du2006improved}
P. Du, W. A. Kibbe and S. M. Lin,
Improved peak detection in mass spectrum by incorporating
continuous wavelet transform-based pattern matching,
\textit{Bioinformatics} 22 (2006) 2059--2065.

\bibitem{stephane1999wavelet}
S. Mallat,
\textit{A Wavelet Tour of Signal Processing} (1999).

\bibitem{noble2006support}
W. S. Noble,
What is a support vector machine?,
\textit{Nature Biotechnology} 24 (2006) 1565--1567.

\bibitem{pisner2020support}
D. A. Pisner and D. M. Schnyer,
Support vector machine,
in: \textit{Machine Learning}, Elsevier (2020), pp. 101--121.

\end{thebibliography}
\end{document}